\documentclass[lettersize,journal]{IEEEtran}
\usepackage{amsmath,amsfonts}
\usepackage{algorithmic}
\usepackage{algorithm}
\usepackage{array}
\usepackage[caption=false,font=normalsize,labelfont=sf,textfont=sf]{subfig}
\usepackage{textcomp}
\usepackage{stfloats}
\usepackage{url}
\usepackage{verbatim}
\usepackage{graphicx}
\usepackage{cite}
\usepackage{multirow}
\usepackage{caption}
\usepackage{xcolor}

\usepackage{color}
\begin{document}

\title{Bidirectional Knowledge Reconfiguration for Lightweight Point Cloud Analysis}

\author{
Peipei Li,
Xing Cui,
Yibo Hu,
Man Zhang,
Ting Yao,~\IEEEmembership{Senior Member,~IEEE,}
and Tao Mei,~\IEEEmembership{Fellow,~IEEE}

\thanks{Peipei Li, Xing Cui and Man Zhang are with the School of Artificial Intelligence, Beijing University of Posts and Telecommunications, Beijing 100876, China. (e-mail: lipeipei@bupt.edu.cn; cuixing@bupt.edu.cn; zhangman@bupt.edu.cn).} 
\thanks{Ting Yao and Tao Mei are with HiDream.ai, Beijing, China. (e-mail: tingyao.ustc@gmail.com; tmei@live.com).} 
\thanks{Corresponding author: Yibo Hu. (e-mail: huyibo871079699@gmail.com).}
\thanks{This work was done during Xing Cui’s internship and Yibo Hu's working in JD AI Research.}
}

\markboth{Journal of \LaTeX\ Class Files,~Vol.~14, No.~8, August~2021}%
{Shell \MakeLowercase{\textit{et al.}}: A Sample Article Using IEEEtran.cls for IEEE Journals}

\IEEEpubid{0000--0000/00\$00.00~\copyright~2021 IEEE}

\maketitle

\begin{abstract}
Point cloud analysis faces computational system overhead, limiting its application on mobile or edge devices. Directly employing small models may result in a significant drop in performance since it is difficult for a small model to adequately capture local structure and global shape information simultaneously, which are essential clues for point cloud analysis. This paper explores feature distillation for lightweight point cloud models. To mitigate the semantic gap between the lightweight student and the cumbersome teacher, we propose bidirectional knowledge reconfiguration (BKR) to distill informative contextual knowledge from the teacher to the student. Specifically, a top-down knowledge reconfiguration and a bottom-up knowledge reconfiguration are developed to inherit diverse local structure information and consistent global shape knowledge from the teacher, respectively.
However, due to the farthest point sampling in most point cloud models, the intermediate features between teacher and student are misaligned, deteriorating the feature distillation performance. To eliminate it, we propose a feature mover's distance (FMD) loss based on optimal transportation, which can measure the distance between unordered point cloud features effectively. Extensive experiments conducted on shape classification, part segmentation, and semantic segmentation benchmarks demonstrate the universality and superiority of our method.
\end{abstract}


\section{Introduction}
\IEEEPARstart{W}{ith} the popularity of 3D sensing devices, 3D data are widely used in many applications, such as autonomous driving, robotics, and virtual reality. Among all kinds of 3D data forms, point clouds are considered a simple but efficient representation. To process irregular, unordered, and unstructured point clouds, early works transform point clouds into regular voxels~\cite{zhou2018voxelnet} or multiview images~\cite{su2015multi}. However, these methods lose rich geometric structure.
Since the success of PointNet~\cite{qi2017pointnet}, processing point clouds directly has been the dominant solution for 3D point cloud analysis~\cite{9760206,9722998}. The subsequent methods, \textit{e.g.}, PointNet++~\cite{qi2017pointnet++}, KCNet~\cite{shen2018mining} and DensePoint~\cite{liu2019densepoint} have achieved significant improvements in point cloud classification and segmentation tasks. These methods can be divided into three categories. 1) MLP-based methods~\cite{qi2017pointnet++,choe2021pointmixer} treat each point independently and map points into high-dimensional features. 2) CNN-based methods~\cite{wu2019pointconv,9410405} design convolution kernels to capture geometric topologies.  3) Transformer-based methods~\cite{zhao2021point,9855233} take advantage of a transformer to extract long-range information.

Despite these advancements, there are still some practical challenges. One is the computational overhead of the system. With the need for applications on mobile or edge devices, point cloud analysis with a small model size, light computation cost, and high performance has attracted much attention. However, current point cloud analysis methods often depend on cumbersome models with expensive computations.
For example, PointTransformer~\cite{zhao2021point} requires more than 18.6 GFLOPs on the ModelNet40 dataset when 1024 points are sampled as input. 
Another key challenge is the irregularity of point clouds, making it difficult to represent discriminative semantic features for elusive shapes. Some methods~\cite{qi2017pointnet,shen2018mining} learn directly from irregular point clouds and sacrifice complexity for effectiveness. Other methods~\cite{qi2017pointnet++,liu2019densepoint} attempt to make full use of the contextual information, including both the global shape and the local structure representations.

\IEEEpubidadjcol
To address the above challenges, in this paper, we investigate lightweight point cloud analysis from the perspective of feature distillation, where the performance of a lightweight student network is improved by transferring informative knowledge from the intermediate features of a cumbersome teacher network. 
As illustrated in Table~\ref{tab:clas}, conventional knowledge distillation algorithms show limited performance in point clouds since the diverse local structure and global shape information of the point cloud are not fully explored during distillation.
To solve this problem, we propose a novel bidirectional knowledge reconfiguration (BKR) mechanism for point cloud feature distillation. 
Specifically, a top-down knowledge reconfiguration and a bottom-up knowledge reconfiguration are designed, where the former is developed for inheriting diverse local structure information from the teacher, and the latter is employed to absorb high-level global shape knowledge from the teacher. 
In addition, we also design a residual connection to encourage distilling knowledge from the same level.
Therefore, BKR mitigates the semantic gap between lightweight students and cumbersome teachers. Additionally, BKR inherits contextual knowledge from the teacher to all the scales of the student. 
In this way, each semantic level of the student network can simultaneously learn contextual information from the teacher network with both the local structure and the global shape knowledge. 

Furthermore, 3D point
clouds are discrete and unordered. Generally, farthest point sampling (FPS) is
employed in most point cloud analysis models~\cite{qi2017pointnet++,wu2019pointconv,zhao2021point} to reduce the resolution of the point cloud.
However, the randomness of FPS results in a misalignment between the intermediate features of the teacher and student, which may further lead to inferior or even destroyed distillation performance.
Inspired by optimal transportation theory~\cite{kusner2015word}, we propose feature mover's distance (FMD) to measure the discrepancy between misaligned teacher and student features.
Specifically, to exploit the local structure information, we divide the transportation task into several subproblems where each subproblem focuses on a local area. 
We further propose a distance-based transportation strategy that approximates the least-expensive transportation flow to simplify the solving procedure of the transportation problem.
Extensive experiments are conducted on several benchmarks, demonstrating the effectiveness and the universality of the proposed method. To summarize, our contributions are fourfold:

\IEEEpubidadjcol 

\begin{itemize}

\item We design a new feature distillation method for lightweight point cloud analysis: a universal knowledge transfer framework for various point cloud models.

\item Bidirectional knowledge reconfiguration (BKR) is proposed to transfer both the low-level structure knowledge and the high-level shape information from the teacher to all the semantic levels of the student. 

\item Since there exists a potential position inconsistency in point cloud features caused by the point sampling operation, the feature mover's distance (FMD) is designed to align the features between the teacher and student.

\item Our method significantly outperforms the previous distillation strategies on point cloud analysis, demonstrating the effectiveness and universality of our framework.

\end{itemize}

\section{Related Work}
In this section, we briefly review existing works related to our method, including point-based classification and segmentation, model compression via knowledge distillation and earth mover's distance. 

\subsection{Point-based Classification and Segmentation}
Methods on point-based classification and segmentation can be divided into three categories: MLP-based~\cite{qi2017pointnet,qi2017pointnet++,choe2021pointmixer}, CNN-based~\cite{wu2019pointconv,9141427,9410405} and transformer-based~\cite{9141427,zhao2021point,wu2022pointconvformer}.
PointNet~\cite{qi2017pointnet} pioneers MLP-based point cloud classification and segmentation. It utilizes MLP to map points to high-dimensional features and aggregates global features through max pooling, thereby extracting permutation invariant features. However, it fails to capture local structures and ignores fine-grained patterns. To solve this problem, PointNet++~\cite{qi2017pointnet++} designs a hierarchical structure to combine features from multiple scales. Although PointNet++ achieves better performance, it still has limitations in information extraction due to the asymmetric structure. To counter this, PointMixer~\cite{choe2021pointmixer} proposes a universal set operator to build a symmetric architecture. Another network, RandLA-Net~\cite{hu2020randla}, improves the efficiency of point cloud processing by using random point sampling instead of point selection.

However, those MLP-based methods only process points individually, ignoring the geometry structure information. To counter this, some researchers intend to design convolution operators on point clouds. DGCNN~\cite{wang2019dynamic} recovers the topological information of the point cloud via a graph and uses Edge-Conv to capture features over a long range. PointConv~\cite{wu2019pointconv} focuses on nonuniform sampling point clouds, a discrete approximation of a continuous convolution. 
To learn relationships in point clouds, DensePoint~\cite{liu2019densepoint} employs relation-shape convolution and builds a dense connection structure to extract dense contextual representations. In contrast, AdaptConv~\cite{zhou2021adaptive} explores an adaptive kernel generated from a pair of points.

More recently, transformer-based methods have been proposed for effective point cloud feature learning. 
Point transformer~\cite{zhao2021point} explores how to extract long-distance relationships in large scenes by developing a self-attention layer for point cloud processing.
Another network, DTNet\cite{9855233}, aggregates pointwise and channelwise self-attention models simultaneously for better feature representation.
Although the above methods show good performance, they all ignore the memory and computational costs. 
\subsection{Model Compression via Knowledge Distillation}
Knowledge distillation~\cite{hinton2015distilling} is a model compression technique that has been widely applied in image processing, such as image classification~\cite{9645162,9834142,9257112}, face analysis~\cite{wu2020learning,she2021dive}, semantic segmentation~\cite{yang2022cross,ji2022structural} and object detection~\cite{qu2022distillation,li2022knowledge}. 
Existing KD methods can be categorized into different categories~\cite{gou2021knowledge}. Based on the number of levels where the distillation occurs, we divide knowledge distillation into single-level methods and multi-level methods. 

For single-level methods, the model distills knowledge only between certain layers of the network. Among them, KD~\cite{hinton2015distilling} minimizes the KL divergence between the last logit outputs of the teacher and the student networks.
Furthermore, DKD~\cite{zhao2022decoupled} improves the flexibility of logit distillation by formulating distillation loss into a target class term and a non-target term.
Recently, many works have focused on optimizing the distillation process via intermediate representations. For example, 
FitNet~\cite{romero2014fitnets} utilizes intermediate features as hints to train a deeper and thinner student. NST~\cite{huang2017like} reviews the distributions of neuron selectivity and matches the distribution between the teacher and student. 
SimKD\cite{chen2022knowledge} designs a simple soft target distillation technique and reuses the classifier layer to narrow the performance gap.
PEFD~\cite{chenimproved} observes the positive effect of the projector in feature distillation.
Therefore, an ensemble of projectors is introduced to improve the performance.

For multi-level methods, knowledge is distilled for multiple layers of the network.
AT~\cite{zagoruyko2016paying} designs several methods for transferring attention maps between the teacher and student.
SP~\cite{tung2019similarity} distills knowledge by preserving the pairwise similarities, which utilizes the pairwise activation similarities within each minibatch to supervise the distillation process.
Recent works, ReviewKD~\cite{chen2021distilling} and SemCKD~\cite{chen2021cross}, further utilize the intermediate features of the teacher model by exploring multilayer knowledge.
Contrary to the aforementioned approaches that are designed for image processing, we introduce knowledge distillation to point cloud analysis aiming at transferring the diverse local structure information and the global shape knowledge from the intermediate point cloud features of a cumbersome teacher to a lightweight student.

\subsection{ Earth Mover's Distance}
\label{sec:related_emd}
Earth mover's distance (EMD)~\cite{rubner2000earth} is proposed to measure the distance between two sets of weighted objects or probability distributions. It has the form of an optimal transportation problem and is defined as the transportation cost under the least-expensive transportation flow. 
Specifically, let $F_{r}=\left \{ \left ( F_{r}^{1},s_{1} \right ),...,\left ( F_{r}^{N},s_{N} \right ) \right \}$ be a set of sources consisting of $N$ pairs, where
${F_{r}^i}$ and ${s_{i}}$ denote the $i\mbox{-}th$ source feature and its corresponding weight, respectively. Let $F_{t}=\left \{ \left ( F_{t}^{1}, t_{1} \right ),..., \left ( F_{t}^{N}, t_{N} \right )\right \}$ be a set of destinations, where
${F_{t}^j}$ and ${t_{j}}$ denote the $j\mbox{-}th$ target feature and its corresponding weight, respectively.
The ground distance between ${F_{r}^i}$ and ${F_{t}^j}$ is denoted by $d_{i,j}$. The goal of the transportation problem is to find the least-expensive flow ${\Pi=\left({\pi_{ij}}\right) \in \mathbb{R}^{N\times N}}$ from ${F_{r}}$ to ${F_{s}}$. The transportation problem can be formulated as a linear programming problem:

\begin{equation}
\label{eq:emd}
\begin{aligned}
&EMD\left ( F_{r},F_{t} \right )=\underset{\Pi \geq 0}{\rm min} \sum_{i,j}d_{i,j}\pi _{i,j}, \\
&
\begin{aligned}
subject\: \: to\: \: 
&\sum_{j} \pi _{i,j} = s_{i},\, \, i\in \left [ 1,N \right ],\\
&\sum_{i} \pi _{i,j} = t_{j},\, \, j\in \left [ 1,N \right ].\\
\end{aligned}
\end{aligned}
\end{equation}

Then, the least-expensive transportation flow can be achieved with the help of linear programming algorithms, such as the Sinkhorn algorithm~\cite{cuturi2013sinkhorn}.

Recently, EMD has been widely used in image processing~\cite{phan2022deepface,doan2022one,zhang2022petsgan}. For example, DeepEMD~\cite{zhang2020deepemd} computes the EMD between dense image features to represent the image distance. DeepFace-EMD~\cite{phan2022deepface} reranks face identification results with EMD to improve out-of-distribution generalization. DensePCR~\cite{mandikal2019dense} predicts the low-resolution point cloud via the EMD loss to measure the consistency of two point sets. 

Despite its effectiveness, EMD is a computationally intensive formulation that requires considerable time and memory.
To alleviate this problem, EXSinkhorn~\cite{chen2022exponential} adds an entropic regularization~\cite{cuturi2013sinkhorn} and adaptively doubles the regularization parameter. SW~\cite{bonneel2015sliced} and its variants~\cite{nguyen2020distributional} characterize high-dimensional probability distributions into one-dimensional space to accelerate the calculation. 
In addition, some works~\cite{fatras2019learning,nguyen2020distributional,nguyen2021transportation} explore minibatch solutions to reduce the memory and computational cost.
BoMb-OT~\cite{nguyen2020distributional} proposes optimal coupling to consider the relationship between minibatches, which approximates the original transportation strategy and constructs a good global mapping. 
Recently, m-POT~\cite{nguyen2021transportation} utilized partial optimal transportation to solve the misspecified mapping problem.

However, these approaches still need to solve complex linear programming problems to find the optimal transportation flow. 
The computational cost is $O\left ( max\left ( m,n \right )^{3} \right )$, which is untenable for the gradient descent-based method. REMD~\cite{kusner2015word,kolkin2019style} solves this problem by relaxing the optimal transportation problem and removing one of the two constraints:

\begin{equation}
R_{F_{r}}\left ( F_{r},F_{t} \right )=\underset{\Pi \geq 0 }{\rm min}\sum_{i,j}d_{i,j}\pi _{i,j}\, \, s.t.\, \, \sum_{j}\pi _{i,j}= s_{i},\\
\end{equation}

\begin{equation}R_{F_{t}}\left ( F_{r},F_{t} \right )=\underset{\Pi \geq 0 }{\rm min}\sum_{i,j}d_{i,j}\pi _{i,j}\, \, s.t.\, \, \sum_{i}\pi _{i,j}= t_{j}.
\end{equation}

Thus, REMD can be formulated as:

\begin{equation}
\label{eq:remd}
    \begin{split}
    L_{REMD}&=REMD\left ( F_{r},F_{t} \right )\\
    &={\rm max}\left (  R_{F_{r}}\left ( F_{r},F_{t} \right ),R_{F_{t}}\left ( F_{r},F_{t} \right )\right )\\
    &={\rm max}( \sum_{i}t_{i}\, \underset{j}{\rm min}\, d_{i,j},\sum_{j}s_{j}\, \underset{i}{\rm min}\, d_{i,j}  ).
    \end{split}   
\end{equation}

Although REMD has shown satisfactory performance in natural language processing~\cite{kusner2015word} and image processing~\cite{kolkin2019style}, it only considers optimal transportation of feature weights globally, failing to transfer local structure information in sparse point clouds effectively. Therefore, we propose a feature mover's distance (FMD) to explore the global shape information as well as the rich local structure information.

\section{Approach}
\subsection{Preliminary}

We denote an input point cloud with $N$ points as $X\in {\mathbb{R}^{N \times d_{in}}}$, where $d_{in}$ is the input dimension. The corresponding positions are defined as $P \in {\mathbb{R}^{N \times 3}}$. 
Usually, $X$ only contains normalized 3D coordinates, \textit{i.e.}, $X=P
$, but it can also be combined with additional attributes, such as surface normal and color.
Given an input $X$ and a lightweight student network $\mathcal{S}$.
The output $Y_s$ can be formulated as:

\begin{equation}
\label{eq:notion1}
\begin{split}
{Y_s} = \mathcal{S}\left(X\right) ={\mathcal{S}_c} \circ{\mathcal{S}_{L}} \circ ... \circ {\mathcal{S}_2} \circ {\mathcal{S}_1} (X),
\end{split}
\end{equation}
where $\mathcal{S}_1,\mathcal{S}_2,...,\mathcal{S}_{L}$ are the sequential blocks of the student. $\mathcal{S}_c$ represents the classifier in the classification task or the decoder in the segmentation task.
$\circ $ is a nesting function, where $g \circ f\left( \cdot  \right) = g\left( {f\left( \cdot \right)} \right)$.
We denote the intermediate features of the student as $\left\{ F_{s,1},F_{s,2},...,F_{s,{L}} \right\}$. $F_{s,l}$ is calculated by:

\begin{equation}
\label{eq:notion2}
F_{s,l} = {{\mathcal{S}_{l}} \circ ... \circ {{\mathcal{S}_2}}\circ {{\mathcal{S}_1}}\left( X \right)}.
\end{equation}

The teacher network $T$ shares a similar process. We denote the intermediate features of the teacher as $\left\{ {{F_{t,1}},{F_{t,2}},...{F_{t,L}}} \right\}$.

\subsection{Overall Framework}

\begin{figure*}
\begin{center}
\includegraphics[width=0.8\linewidth]{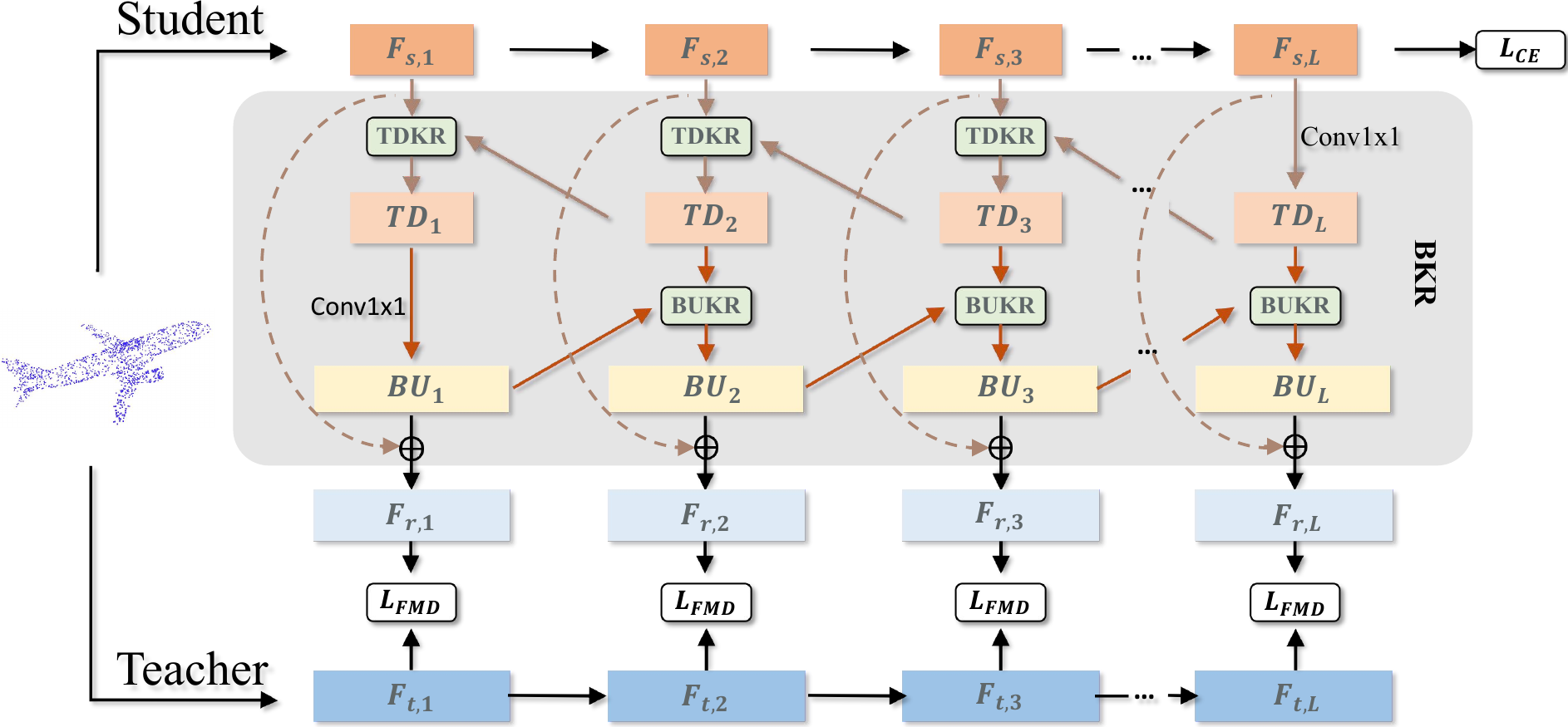}
\end{center}
\caption{
The overall framework of our method. Bidirectional knowledge reconfiguration (BKR) contains top-down knowledge reconfiguration (TDKR), bottom-up knowledge reconfiguration (BUKR), and residual connection.
}
\label{fig:framework_my}

\end{figure*}

In this paper, we propose bidirectional knowledge reconfiguration (BKR), a novel feature distillation mechanism, for lightweight point cloud analysis. The overall framework is illustrated in Fig.~\ref{fig:framework_my}. 
BKR consists of top-down knowledge reconfiguration (TDKR), bottom-up knowledge reconfiguration (BUKR) and residual connection (RES), aiming at alleviating the semantic gap between teacher and student, as well as distilling the contextual knowledge from the teacher to all the semantic levels of the student. 
However, we observe that position inconsistency between the corresponding teacher and student features, caused by the random sampling operation, is one of the main factors affecting the performance of point cloud feature distillation. To solve this problem, we further design a feature mover's distance (FMD), which can measure the discrepancy between misaligned student features and teacher features effectively.

\subsection{Bidirectional Knowledge Reconfiguration}

Multi-level distillation is widely employed in feature distillation and shows satisfactory performance~\cite{zagoruyko2016paying,yim2017gift,tung2019similarity}, which usually transfers the same-level knowledge between teacher and student.
However, in point cloud analysis, neglecting cross-level knowledge may lead to a loss of rich 3D geometric information and is not conducive to grasping the diverse shape information formed by point clouds~\cite{liu2019densepoint}.  
Inspired by multiscale feature learning~\cite{guo2020augfpn,liu2018path}, we propose bidirectional knowledge reconfiguration for point cloud feature distillation, imposing multi-level and multiscale contextual knowledge from the teacher to all the semantic levels of the student hierarchically. 
We divide layers with the same resolution into a group and view them as a level. In each level, the feature of the last layer is employed to distill the knowledge.
Specifically, a top-down knowledge reconfiguration is first employed to merge the information from top to bottom of the student so that the low-level structure knowledge of the teacher can be spread to deep student layers.
In addition to low-level structure knowledge, features at high levels represent global knowledge, which is essential for perceiving the overall shape of the point cloud. To further inherit the high-level shape knowledge from the teacher, we perform a bottom-up knowledge reconfiguration on the features produced by the top-down knowledge reconfiguration. Finally, the reconfigured feature and the original student feature are fused via a residual connection to better inherent information from the same level.

\begin{figure*}
\begin{center}
\includegraphics[width=\linewidth]{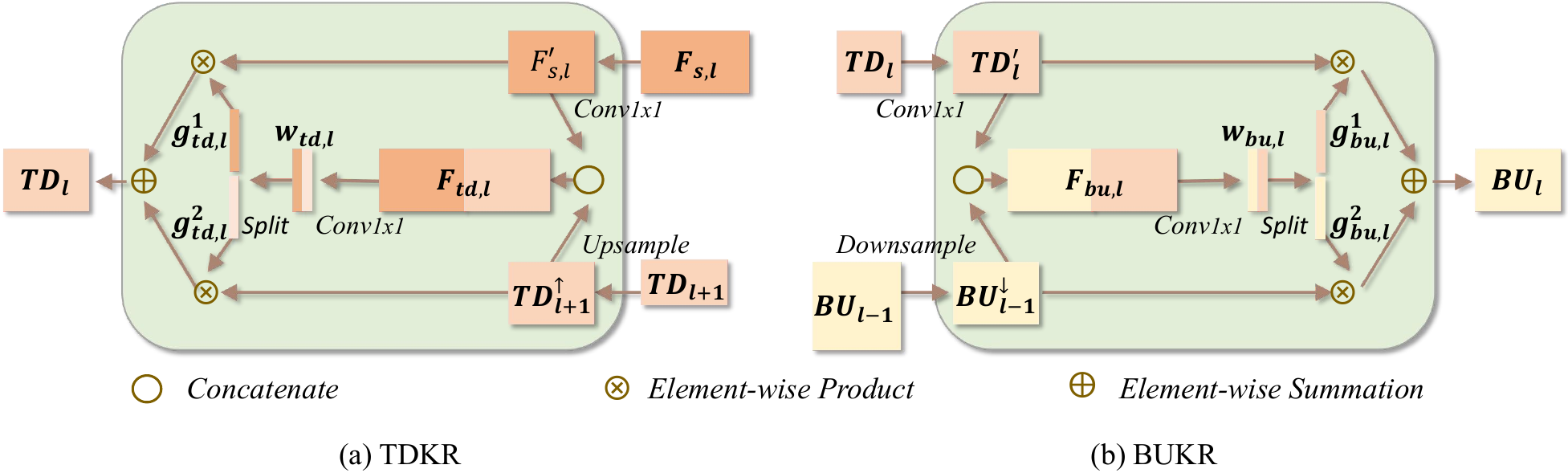}
\end{center}
   \caption{The architectures of (a) top-down knowledge reconfiguration (TDKR) and (b) bottom-up knowledge reconfiguration (BUKR). 
   The size of the feature blocks represents the relative shape of the features.}
\label{fig:block}
\end{figure*}

\subsubsection{Top-down Knowledge Reconfiguration (TDKR)}

As shown in Fig.~\ref{fig:framework_my}, we denote {$\left\{ {{TD_1},{TD_2},...,{TD_{L}}} \right\}$} as the reconfigured features of TDKR, where ${TD_ {l}}$ is formulated as:

\begin{equation}
\label{eq:tdkr}
T{D_{l}} = \left\{ \begin{array}{ll}
TDKR\left( {T{D_{{l} + 1}},F_{s,{l}}} \right),\;\;  &{l = 1,...,L - 1}\\
{Conv1\!\times\!1\left(F_{s,{l}}\right)},  &{l = L} \\
\end{array}  \right. .
\end{equation}

Fig.~\ref{fig:block}(a) presents the building block of TDKR. 
Taking the $l\mbox{-}th$ level of the student as an example, with the reconfigured feature $TD_{{l}+1} \in \mathbb{R}^{{n'}\times d}$ from the $\left( {l+1} \right) \mbox{-}th$ level, we first upsample the feature resolution to the same size as the corresponding teacher feature. Similar to~\cite{qi2017pointnet++}, we obtain the upsampled feature by interpolating feature values of $(l+1)\mbox{-}th$ level points at coordinates of the $l\mbox{-}th$ level points.
The output is denoted as ${{TD_{{l} + 1}}^ \uparrow } \in \mathbb{R}^{{n}\times d}$:

\begin{equation}
\label{eq:tdi_}
TD_{l + 1}^ \uparrow  = Upsample\left( {T{D_{l + 1}}} \right).
\end{equation}

Specifically, if the feature is global, we simply use repetition as the upsampling operation.
Additionally, the original student feature {${F_{s,{l}}} \in {\mathbb{R}^{n \times d'}}$} undergoes a $1 \times 1$ convolution to match the dimension of {$TD_{l + 1}^ \uparrow $}, which is termed {$F_{s,{l}}^{'} \in {\mathbb{R}^{n \times d}}$}:

\begin{equation}
\begin{array}{ll}
\label{eq:fsi_}
F{_{s,{l}}^{'}} = Conv1 \times 1\left( {{F_{s,{l}}}} \right).
\end{array}\end{equation}

Inspired by~\cite{li2019selective,hu2018squeeze}, we employ a gate mechanism to control the information flows from different features. Specifically, we concatenate {${TD_{{{l} + 1}}^ \uparrow }$} and {${F_{s,{l}}^{'}}$} as {${F_{_{td,{l}}}} \in \mathbb{R}^{n \times 2d}$} and employ a 1$\times$1 convolution with a sigmoid function to generate the weight {$w_{_{td,{l}}} \in \mathbb{R}^{n \times 2}$}. Then, the weight is split into two gates {$g_{_{td,{l}}}^1 \in \mathbb{R}^{n \times 1}$} and {$g_{_{td,{l}}}^2 \in \mathbb{R}^{n \times 1}$}. TDKR is calculated as:

\begin{equation}\label{eq:tdkr_detail}
{TDKR\left( {T{D_{{l} + 1}},{F_{s,{l}}}} \right) = {[g_{_{td},{l}}^{1}]} \cdot {F_{s,{l}}^{'}}} + {[g_{_{td},{l}}^{2}]} \cdot {TD_{{l} + 1}^{\uparrow }},
\end{equation}
where $[g_{_{td},l}^{1}] \in \mathbb{R}^{n\times d}$ and $[g_{_{td},l}^{2}] \in \mathbb{R}^{n\times d}$ are the repetitions of $g_{_{td},l}^{1}$ and $g_{_{td},l}^{2}$ $d$ times, respectively.
In this way, the weights are generated dynamically based on the input features. Thus, the information flows from different levels that carry diverse knowledge can be reconfigured adaptively.

\subsubsection{Bottom-up Knowledge Reconfiguration (BUKR)} As illustrated in Fig.~\ref{fig:block} (b), the structure of BUKR is similar to that of TDKR but different in detail: BUKR performs downsampling on low-level features for feature fusion, while TDKR performs upsampling on high-level features. We define $\left\{ {B{U_1},B{U_2},..., \\B{U_{L}}}\right\}$ as the
outputs of BUKR, where {${BU_{l}}$} is formulated as:

\begin{equation}
\label{eq:bukr}
{BU_{l}} = \left\{ \begin{array}{l}
BUKR\left( {B{U_{l - 1}},{TD_{l}}} \right),\quad l = 2,...,L\\
Conv1\!\times\!1\left(TD_{l}\right),\quad\quad\quad\; l = 1
\end{array} \right. .
\end{equation}

Specifically, ${BU_{l - 1}}$ is first downsampled to match the resolution, and a $1\times1$ convolution is performed on $TD_{l}$ to match the dimension:

\begin{equation}
\begin{array}{l}
    {BU_{l - 1}^ \downarrow}=Downsample\left( {B{U_{l - 1}}} \right) ,
    \end{array}
\end{equation}
\begin{equation}
    {TD_{l}^{'}} = Conv1 \times 1\left( {{TD_{l}}} \right).
\end{equation}

Then, we calculate the weight $w_{_{bu,l}} \in \mathbb{R}^{n \times 2}$ from the concatenation of ${BU_{l - 1}^ \downarrow}$ and ${TD_{l}^{'}}$ in the same way as TDKR. $w_{_{bu,l}} $ is further split into $g_{_{bu,l}}^1 \in \mathbb{R}^{n \times 1}$ and $g_{_{bu,l}}^2 \in \mathbb{R}^{n \times 1}$ as two gates. Finally, the output of BUKR can be written as:

\begin{equation}\label{eq:bukr_detail}
	BUKR\left( {B{U_{l - 1}},{TD_{l}}} \right)=  {[g_{_{bu,l}}^{1}]} \cdot {TD_{l}^{'}} + {[g_{_{bu,l}}^{2}]} \cdot {BU_{l - 1}^{\downarrow}},
\end{equation}
where the notations are similar to TDKR.

\begin{figure}
\begin{center}
\includegraphics[width=0.8\linewidth]{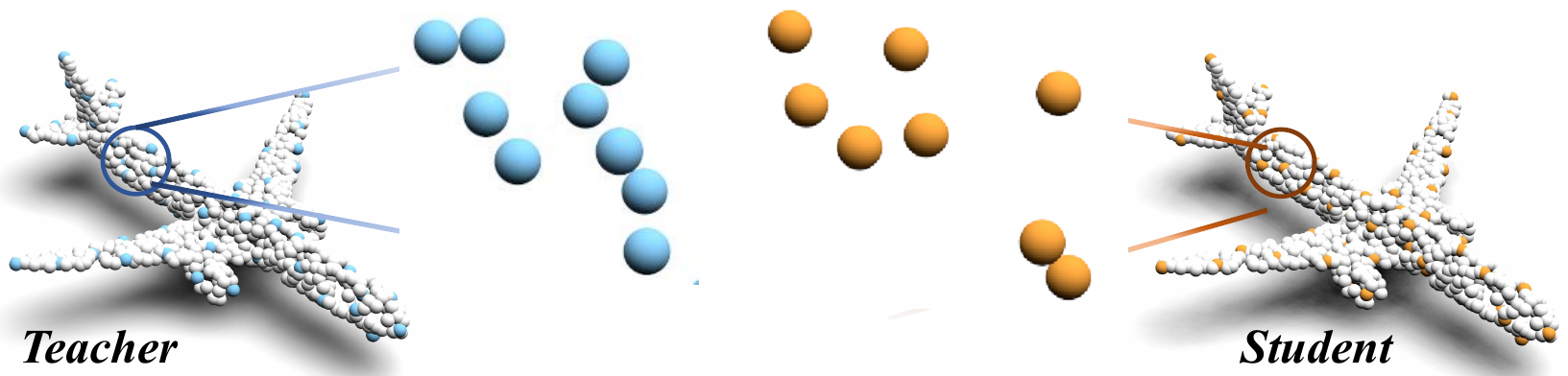}
\end{center}
   \caption{Illustration of position inconsistency. 
   The blue dots represent teacher features. The orange dots are student features.
   }\label{fig:qsc}
\end{figure}

\subsubsection{Residual Connection (RES)} TDKR and BUKR can transfer cross-level information, while rich knowledge at the same level might be ignored. To effectively inherit the same level of knowledge from the teacher, a residual connection is employed to obtain the reconfigured feature $F_{r,l}$:

\begin{equation}
    F_{r,l}=BU_{l}+F_{s,l}.
\end{equation}

\subsection{Feature Mover's Distance}
As shown in Fig.~\ref{fig:qsc}, 
The randomness of farthest point sampling (FPS) makes the position and order of points different between the teacher and the student.
Taking the $l\mbox{-}th$ level as an example, after FPS, the point positions of the student ${P_{s,l}} \in {R^{N \times 3}}$ are not equal to the point positions of the teacher ${P_{t,l}} \in {R^{N \times 3}}$, leading to feature misalignment between the teacher and student.
We present more analysis in Section~\ref{sec:inconsistency} to show the misalignment.
To this end, it is essential to align the point positions of the intermediate features before distilling the knowledge from teacher to student.

Inspired by the optimal transportation theory, we propose the feature mover's distance (FMD) to align the point positions of the features between the teacher and student. Specifically, we first divide the original optimal transportation problem into $N$ subproblems to leverage local structure information. Denote $F_{r,l}=\left \{ F_{r,l}^{1}, ...,F_{r,l}^{N} \right \}$ as the reconfigured feature of the student and $P_{s,l}=\left \{ P_{s,l}^{1}, ...,P_{s,l}^{N} \right \}$ as its corresponding positions. We divide the student feature into $N$ subsets. Therefore, each subset contains one element, \textit{i.e.}, $\hat{{F}_{r,l}^{i}}=\left \{{F}_{r,l}^{j}  \mid j=i \right \}$. Similarly, let $F_{t,l}$ be the feature of the teacher and $P_{t,l}=\left \{ P_{t,l}^{1}, ...P_{t,l}^{N} \right \}$ be its corresponding positions. 
As transporting products to neighboring destinations is an approximation of the least-expensive transportation strategy~\cite{ricci2012prototype,wagner2010efficient}, we define the teacher feature subset $\hat{F_{t,l}^{i}}=\left \{{F}_{t,l}^{j} \mid j \in {N_{P_{s,l}}^i}\left( {{P_{t,l}}} \right) \right \}$.
${N_{P_{s,l}}^i}\left( {{P_{t,l}}} \right)$ is the index of the $k$ nearest neighbors of student position $P_{s,l}^i$ in teacher position set $P_{t,l}$.

Finally, we define FMD as the feature discrepancy under a distance-based transportation strategy.
Specifically, we propose a distance-based transportation strategy $\Pi^{l}=\left({\pi^{l}}_{ij}\right)$ to approximate the least-expensive transportation strategy. Similar to~\cite{lang2020samplenet}, we determine the transportation strategy based on the ground distance $ d_{i,j}$ and use the normalized Gaussian radial basis function to calculate the $\pi^{l}_{i,j}$ of the $l\mbox{-}th$ level features:

\begin{equation}
\label{eq:pi}
    \pi^{l}_{i,j}=\frac{e^{-d_{i,j}^{2}/2\tau ^{2}}}{\sum_{h\in {N_{P_{s,l}^i}}\left( {{P_{t,l}}} \right)}e^{-d_{i,h}^{2}/2\tau ^{2}}},
\end{equation}
where $d_{i,j}=\left \| P_{s,l}^{i}-P_{t,l}^{j} \right \|_{2}$ and $\tau$ is a temperature parameter. FMD is calculated as follows: 

\begin{equation}
\label{eq:pfd}
\begin{split}
    L_{FMD}^{l}&=FMD\left ( F_{r,l},F_{t,l} \right )\\
    &=\sum_{i=1}^{N}s_{i,l}\left \| F_{r,l}^{i}-\sum_{j \in {N_{P_{s,l}^i}}\left( {{P_{t,l}}} \right)}\pi^{l}_{i,j}F_{t,l}^{j} \right \|_{2},
\end{split}
\end{equation}
where $F_{r,l}$ and $F_{t,l}$ are the reconfigured student feature and the teacher feature of the $l\mbox{-}th$ level, respectively. 
Compared to REMD in Eq. (\ref{eq:remd}) that only considers the global nearest destination, FMD takes $k$ nearest local neighbors into account, which transfers local structure information of different levels effectively and makes the measurement more robust.

Inspired by average pooling correlation (APC)~\cite{phan2022deepface}, we formulate $s_{i,l}$ as follows:

\begin{equation}
    s_{i,l}=max(0, \langle F_{r,l}^{i},\frac{\sum_{j}^{N}F_{t,l}^{j}}{N}  \rangle).
\end{equation}

During the training process, we utilize both the original cross-entropy loss $ {L_{CE}}$ and the FMD loss $ L_{FMD}^{l}$. The total loss function is:
\begin{equation}
\label{eq:loss_total}
L = {L_{CE}} + \lambda {\sum_{l=1}^{L}L_{FMD}^{l}},
\end{equation}
where $\lambda$ is a trade-off hyperparameter.

\begin{table}
\scriptsize
\renewcommand\arraystretch{1.5}
\centering
\caption{Resource usage for different models. 
T and S represent the original teacher and compressed student model.}
\setlength{\tabcolsep}{1.3mm}
\scalebox{0.89}{
\begin{tabular}{ll|cc|cc|cc}
\hline
&  &\multicolumn{2}{c|}{\textbf{Shape Classification}}&\multicolumn{2}{c|}{\textbf{Object Part Segmentation}}&\multicolumn{2}{c}{\textbf{Semantic Segmentation}}\\
\cline{3-8}

&&MAdds(M)&Params(M)&MAdds(M)&Params(M)&MAdds(M)&Params(M)\\
\cline{1-8}
\multirow{2}{*}{\textbf{PN++}} &T  &868 &1.48 &1154 &1.41 &1042 &0.97 \\
   &S  &19 &0.03 &35 &0.03 &62 &0.02\\
\cline{1-8}
\multirow{2}{*}{\textbf{DGC}} &T  &2449 &1.81 &4538 &1.46 &6181 &0.98\\
   &S  &45 &0.03 &1149 &0.82 &127 &0.02\\
\cline{1-8}
\multirow{2}{*}{\textbf{PConv}} &T  &1171 &19.57 &10012 &10.12 &9990 &10.11\\
   &S  &41 &0.31 &396 &0.17 &247 &0.17\\
\cline{1-8}
\multirow{2}{*}{\textbf{PT}} &T  &18600  &9.58 &37840 &19.40 &/ &/\\
   &S  &320 &0.15 &740 &1.92 &/ &/\\
\hline

\end{tabular}
}
\label{tab:flops}
\end{table}

\section{Experiments}
We evaluate the effectiveness of our method on ModelNet40~\cite{wu20153d} for point cloud classification, ShapeNetPart~\cite{yi2016scalable} for object part segmentation and S3DIS~\cite{armeni20163d} for point cloud semantic segmentation.
Since there are few studies on point cloud distillation, we select the Feature-L2 (F-L2) as our baseline, which is a classical feature distillation method in image processing.
In F-L2, all the intermediate features of the student are first transformed to match the size of the corresponding teacher features. Then, $L_2$ loss is employed as the distillation objective.
We choose widely used distillation methods as competitors, including KD~\cite{hinton2015distilling}, FitNet~\cite{romero2014fitnets}, NST~\cite{huang2017like}, AT~\cite{zagoruyko2016paying}, SP~\cite{tung2019similarity}, OFD~\cite{heo2019comprehensive}, DKD~\cite{zhao2022decoupled} and PEFD~\cite{chenimproved}. Four classical models are chosen as the backbones, including the MLP-based model PointNet++ (PN++)~\cite{qi2017pointnet++}, graph-based model DGCNN (DGC)~\cite{wang2019dynamic}, CNN-based model PointConv (PConv)~\cite{wu2019pointconv} and transformer-based model PointTransformer (PT)~\cite{zhao2021point}. 
We treat the original model as the teacher and reduce the width to 1/8 as the student, which is marked by a prefix of (1/8). Table~\ref{tab:flops} lists the number of parameters (Params) and the multiadds (MAdds) of these models.

For PointNet++~\cite{qi2017pointnet++} and PointConv~\cite{wu2019pointconv}, in addition to the point positions, we also employ the surface normals as the additional input. 
For all the experiments, we set the data augmentations and the training hyperparameters the same as 
the open source codes$\footnote{\url{ https://github.com/yanx27/Pointnet_Pointnet2_pytorch}}$$^{,}\footnote{\url{ https://github.com/AnTao97/dgcnn.pytorch}}$$^{,}\footnote{\url{ https://github.com/DylanWusee/pointconv_pytorch}}$$^{,}\footnote{\url{https://github.com/qq456cvb/Point-Transformers}}$. For the tradeoff parameter $\lambda$, we conduct cross-validation on the ModelNet40 dataset and find that $\lambda=0.1$ achieves the best results for classification and $\lambda=0.01$ achieves the best results for segmentation. For the competitors, we also conduct the same cross-validation experiment to choose the tradeoff parameter. In addition, we choose $k=5$ as the number of neighbors in FMD.
Code is available at \textit{https://github.com/cuixing100876/BKR}.

\begin{table} 
\scriptsize
\renewcommand\arraystretch{1.5}
\centering\caption{Results on shape classification.
}
\setlength{\tabcolsep}{0.8mm}
\scalebox{0.85}{
\begin{tabular}{l|cc|ccccccccc|c}

\hline
\cline{1-13}
& \textbf{PN++}&\textbf{(1/8)PN++} &\textbf{F-L2}&\textbf{SP}&\textbf{FitNet}&\textbf{NST}&\textbf{AT}&\textbf{KD}&\textbf{OFD}&\textbf{DKD}&\textbf{PEFD}&\textbf{Ours}\\
\hline
\textbf{OA}& 92.54 & 88.48 & 88.03 & 88.31 & 88.57 & 88.61 & 89.08 & 89.02 & 89.23 & 89.18 &89.28 &\textbf{90.28} \\
\textbf{mAcc}& 90.17 & 80.61 & 79.69 & 80.03 & 82.17 & 82.59 & 82.79 & 82.94  & {83.25}  & 83.08 & 83.27 &\textbf{84.55} \\

\hline
&\textbf{DGC}&\textbf{(1/8)DGC}  &\textbf{F-L2}&\textbf{SP}&\textbf{FitNet}&\textbf{NST}&\textbf{AT}&\textbf{KD}&\textbf{OFD}&\textbf{DKD}&\textbf{PEFD}&\textbf{Ours}\\
 
\hline
\textbf{OA}& 92.26 & 82.65 & 84.03 & 84.62 & 83.72 & 84.22 & 84.24 & 83.60   & {85.04} & 83.63 & 85.27 &\textbf{86.46} \\
\textbf{mAcc}& 89.42 & 68.09 & 71.19 & 72.25 & 71.13 & 71.55 & 71.95 & 69.76   & {72.40} & 69.89 & 72.71 &\textbf{74.77} \\

\hline
& \textbf{PConv}&\textbf{(1/8)PConv} &\textbf{F-L2}&\textbf{SP}&\textbf{FitNet}&\textbf{NST}&\textbf{AT}&\textbf{KD}&\textbf{OFD}&\textbf{DKD}&\textbf{PEFD}&\textbf{Ours}\\

\hline
\textbf{OA}& 92.34 & 74.53 & 73.23 & 74.44 & 74.07 & 74.38 & 76.38 & 76.74   & {77.32} & 76.96 & 78.13 &\textbf{83.73} \\
\textbf{mAcc}& 89.15 & 62.38 & 60.43 & 60.57 & 60.75 & 61.01 & 62.74 & 63.81   & {65.15} & 64.03 & 67.51 &\textbf{72.16} \\

\hline
& \textbf{PT}&\textbf{(1/8)PT} &\textbf{F-L2}&\textbf{SP}&\textbf{FitNet}&\textbf{NST}&\textbf{AT}&\textbf{KD}&\textbf{OFD}&\textbf{DKD}&\textbf{PEFD}&\textbf{Ours}\\

\hline
\textbf{OA}& 92.31 & 87.18 & 86.57 & 86.69 & 87.42 & 87.66 & 87.82 & 87.90   & {86.01} & 87.88 & 87.98 &\textbf{88.50} \\
\textbf{mAcc}  & 89.92 & 82.58 & 80.83 & 81.49 & 82.63 & 82.99 & 83.47 & 83.00   & {80.73} & 83.11 & 83.89 &\textbf{84.25} \\

\hline
\cline{1-13}
\end{tabular}
}
\label{tab:clas}
\end{table}

\subsection{Shape Classification}
\subsubsection{Data and Metrics} The ModelNet40 dataset~\cite{wu20153d} consists of 12,311 meshed CAD models from 40 categories, which is split into 9,843 models for training and 2,468 models for testing. We follow the data preparation of~\cite{wu20153d} and employ the mean accuracy within each category (mAcc) and the overall accuracy (OA) as the evaluation metrics.

\subsubsection{Results} 
As shown in Table~\ref{tab:clas}, when dealing with models that involve sampling operations, such as PointNet++, PointConv, and PointTransformer, the utilization of F-L2 may have a negative impact on the performance of student model. This is because directly transferring intermediate feature knowledge without reconfiguration and alignment may potentially result in performance degradation. 
Our method (BKR+FMD) outperforms other distillation methods on all four backbones.
Specifically, for the PointNet++ model, our method outperforms DKD by 1.10\% and 1.47\% in OA and mAcc, respectively. Moreover, our method also outperforms PEFD by large margins, \textit{i.e.}, 1.00\% and 1.28\% in OA and mAcc, respectively, which demonstrates the effectiveness of BKR and FMD.
Besides, our method improves the mACC of the student by 3.94\%, 6.68\%, 9.78\% and 1.67\% and the OA by 1.8\%, 3.81\%, 9.2\% and 1.32\% with PointNet++, DGCNN, PointConv and PointTransformer, respectively.
The improvement demonstrates that the proposed BKR and FMD can benefit the knowledge transfer procedure in various kinds of point cloud models, demonstrating the universality of our method.

\begin{table}
\scriptsize
\renewcommand\arraystretch{1.5}
\centering
\caption{Results on object part segmentation.
}
\setlength{\tabcolsep}{0.8mm}
\scalebox{0.82}{
\begin{tabular}{l|cc|ccccccccc|c}

\cline{1-13}
\hline
& \textbf{PN++}&\textbf{(1/8)PN++}&\textbf{F-L2}&\textbf{SP}&\textbf{FitNet}&\textbf{NST}&\textbf{AT}&\textbf{KD}&\textbf{OFD}&\textbf{DKD}&\textbf{PEFD}&\textbf{Ours}\\

\hline
\textbf{ins.mIoU}& 85.21 & 76.29 & 75.92 & 75.99 & 76.34 & 76.50 & 76.71 & 76.82 & 76.94 &76.84 &77.25 &\textbf{79.22} \\
\textbf{cat.mIoU}& 81.74 & 58.05 & 57.93 & 57.97 & 58.03 & 58.37 & 58.25 & 58.32 & 58.60 &58.37 &58.43 &\textbf{59.84} \\

\hline
 &\textbf{DGC}&\textbf{(1/8)DGC}&\textbf{F-L2}&\textbf{SP}&\textbf{FitNet}&\textbf{NST}&\textbf{AT}&\textbf{KD}&\textbf{OFD}&\textbf{DKD}&\textbf{PEFD}&\textbf{Ours}\\

\hline
\textbf{ins.mIoU}& 84.86 & 72.39 & 73.02 & 73.57 & 72.95 & 73.07 & 73.44 & 72.54 & 74.24 &72.95 &74.64  &\textbf{76.36} \\
\textbf{cat.mIoU}& 82.23 & 48.75 & 51.55 & 55.56 & 51.24 & 51.63 & 55.32 & 50.64  & 55.85 &50.71 &55.98  &\textbf{57.53} \\

\hline
& \textbf{PConv}&\textbf{(1/8)PConv}&\textbf{F-L2}&\textbf{SP}&\textbf{FitNet}&\textbf{NST}&\textbf{AT}&\textbf{KD}&\textbf{OFD}&\textbf{DKD}&\textbf{PEFD}&\textbf{Ours}\\

\hline
\textbf{ins.mIoU} & 85.18 & 78.95 & 76.36 & 76.50 & 77.01 & 78.37 & 79.65 & 79.91 & 80.02 &79.90 &79.93  &\textbf{80.22} \\
\textbf{cat.mIoU}& 81.95 & 59.98 & 55.83 & 55.85 & 56.73 & 59.87 & 60.90 & 61.47  & 61.67 &61.55 &62.01  &\textbf{63.38} \\

\hline
& \textbf{PT}&\textbf{(1/8)PT} &\textbf{F-L2}&\textbf{SP}&\textbf{FitNet}&\textbf{NST}&\textbf{AT}&\textbf{KD}&\textbf{OFD}&\textbf{DKD}&\textbf{PEFD}&\textbf{Ours}\\

\hline
\textbf{ins.mIoU} & 83.75 & 74.76 & 73.57 & 75.06 & 75.14 & 75.30 & 75.50 & 75.90  & 74.32 &75.76 &75.84  &\textbf{77.83}\\
\textbf{cat.mIoU} & 79.95 & 60.76 & 58.47 & 64.17 & 64.56 & 64.61 & 65.00 & 65.56  & 58.51 &65.44 &65.65  &\textbf{66.15} \\

\hline
\cline{1-13}
\end{tabular}
}
\label{tab:part}
\end{table}

\subsection{Object Part Segmentation}

\subsubsection{Data and Metrics} The ShapeNetPart dataset~\cite{yi2016scalable} contains 16,880 models from 16 shape classes. There are 14,006 models for training and 2,874 models for testing. Each point is annotated with one label from 50 parts, and the number of parts for each class is 2-6. For a fair comparison, we follow the same testing protocol with~\cite{wu20153d}. The category mIoU and the instance mIoU are employed for evaluation.

\subsubsection{Results} Similar to the experiments on ModelNet40, we compare our method (BKR+FMD) with the competitors on all four backbones for the object part segmentation task. The results are presented in Table~\ref{tab:part}. 
Our method surpasses PEFD by 1.97\% and 1.41\% on instance mIoU and category mIoU for PointNet++~\cite{qi2017pointnet++}, respectively.
In the case of DGCNN, although all the distillation methods improve the performance of the original student, our method achieves the largest improvement. 
For PointConv, our method achieves an improvement of $1.27\%$ and $3.4\%$ on instance mIoU and category mIoU, respectively.
The success on the object part segmentation task further reveals the applicability of our method.

\begin{table}
\scriptsize
\renewcommand\arraystretch{1.5}
\centering
\caption{Results on semantic segmentation.
}
\setlength{\tabcolsep}{0.8mm}
\scalebox{0.82}{
\begin{tabular}{l|cc|ccccccccc|c}

\hline
\cline{1-13}
& \textbf{PN++}&\textbf{(1/8)PN++} &\textbf{F-L2}&\textbf{SP}&\textbf{FitNet}&\textbf{NST}&\textbf{AT}&\textbf{KD}&\textbf{OFD}&\textbf{DKD}&\textbf{PEFD}&\textbf{Ours}\\

\hline
\textbf{OA}& 82.75 & 79.58 & 79.26 & 79.38 & 79.68 & 79.69 & 80.26 & 80.33 & 80.58 &80.40 &80.61 &\textbf{81.02} \\
\textbf{mAcc}& 61.16 & 57.25 & 56.89 & 57.20 & 57.75 & 57.63 & 58.01 & 58.17 & 58.60 &58.47 &58.73 &\textbf{60.30} \\
\textbf{mIoU}& 52.23 & 46.09 & 45.34 & 45.83 & 46.39 & 46.24 & 47.27 & 47.42 & 48.32 &48.15 &48.44 &\textbf{50.03} \\

\hline
&\textbf{DGC}&\textbf{(1/8)DGC} &\textbf{F-L2}&\textbf{SP}&\textbf{FitNet}&\textbf{NST}&\textbf{AT}&\textbf{KD}&\textbf{OFD}&\textbf{DKD}&\textbf{PEFD}&\textbf{Ours}\\

\hline
\textbf{OA}& 83.70 & 77.19 & 78.59 & 79.39 & 78.24 & 78.72 & 79.16 & 77.83 & 79.44 &78.36 &79.56 &\textbf{80.05} \\
\textbf{mAcc}& 54.07 & 43.16 & 46.16 & 47.49 & 44.17 & 47.02 & 47.14 & 44.03 & 47.77 &44.52 &47.83 &\textbf{48.50} \\
\textbf{mIoU}& 47.21 & 35.51 & 36.78 & 39.62 & 36.50 & 37.28 & 38.42 & 35.92 & 39.26 &36.69 &39.19 &\textbf{39.92} \\

\hline
& \textbf{PConv}&\textbf{(1/8)PConv} &\textbf{F-L2}&\textbf{SP}&\textbf{FitNet}&\textbf{NST}&\textbf{AT}&\textbf{KD}&\textbf{OFD}&\textbf{DKD}&\textbf{PEFD}&\textbf{Ours}\\

\hline
\textbf{OA}& 84.76 & 81.99 & 81.57 & 81.56 & 81.76 & 81.84 & 82.29 & 82.33 & 82.58 &82.47 &82.68 &\textbf{83.62} \\
\textbf{mAcc}& 65.68 & 60.04 & 59.57 & 59.64 & 59.78 & 59.95 & 61.34 & 61.56 & 61.83 &61.55 &61.78 &\textbf{62.52} \\
\textbf{mIoU}& 55.31 & 51.47 & 50.84 & 51.04 & 51.10 & 51.23 & 51.45 & 51.77 & 52.23 &51.84 &52.14 &\textbf{52.75} \\
\hline
\cline{1-13}
\end{tabular}
}
\label{tab:semantic}
\end{table}

\begin{figure}
\begin{center}
\includegraphics[width=0.95\linewidth]{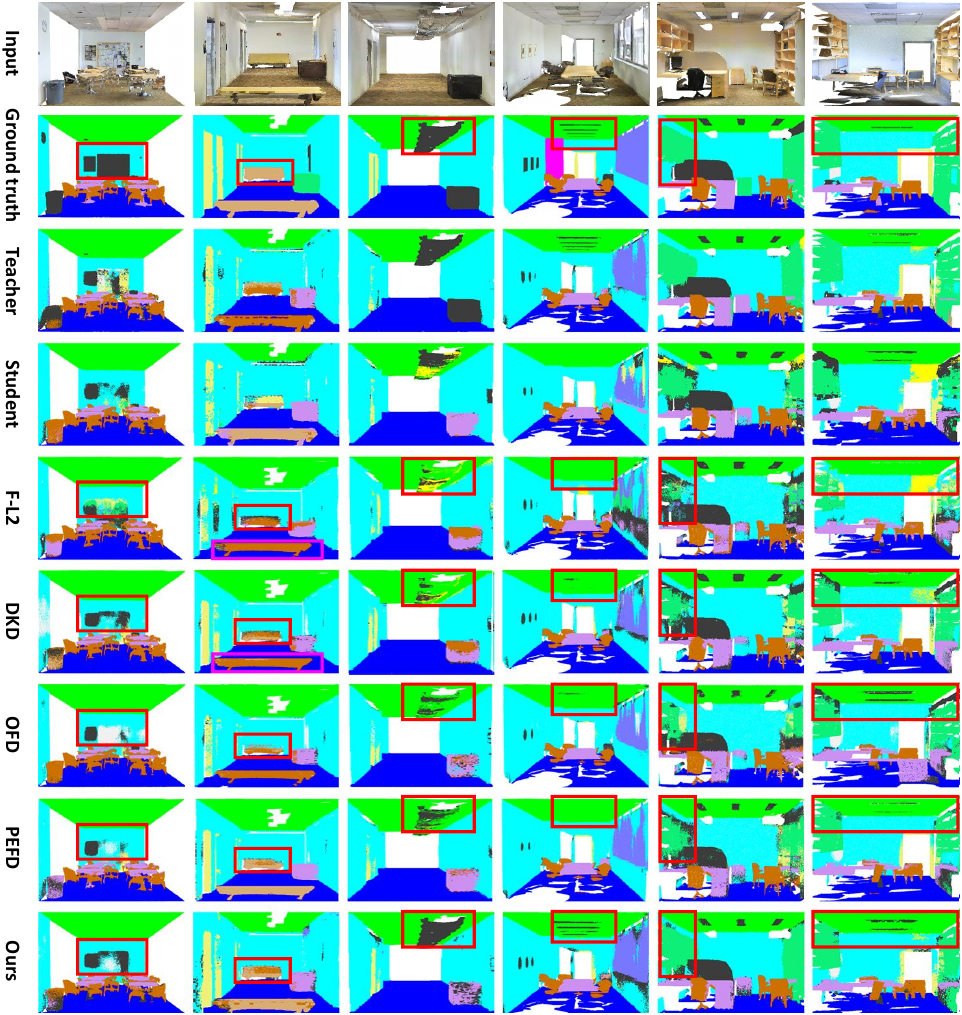}
\end{center}
   \caption{Visualization of semantic segmentation results.
   }\label{fig:sem}
\end{figure}

\subsection{Semantic Segmentation}

\subsubsection{Data and Metrics} The S3DIS dataset~\cite{armeni20163d} contains 271 rooms in 6 indoor areas. There are 273 million 3D RGB points scanned from three different buildings, each of which is assigned a semantic label from 13 classes. We train the models on Areas 1-4 and 6 and test on Area 5, which is unseen during training. The mean classwise intersection over union (mIoU), mAcc and OA are employed as the evaluation metrics.

\subsubsection{Results}
Similarly, we conduct comparisons between our method and other distillation methods on several backbones. However, due to the more complex scenarios and serious self-obscuring, semantic segmentation is more challenging than object part segmentation, leading to less effectiveness of the distillation. As shown in Table~\ref{tab:semantic}, there are performance drops or slight performance improvements with 
previous distillation methods. With the help of reconfiguration and alignment, our method consistently and significantly improves the semantic segmentation performance, especially on mAcc and mIoU.

\subsubsection{Visualization}
Fig.~\ref{fig:sem} presents the visualization results of PointNet++~\cite{qi2017pointnet++} on S3DIS. The predictions of our method are closer to the ground truth and capture more connected and consistent local details since the bidirectional knowledge reconfiguration has the ability to well inherit the contextual knowledge from the teacher model. Besides, Fig.~\ref{fig:sem} also shows qualitative comparisons between our method with other competing methods, including F-L2, DKD, OFD and PEFD.
As shown in Fig.~\ref{fig:sem}, F-L2 obtains unsatisfactory performance, which even degenerates the performance of the student model. This may be primarily due to its inherent problem of position inconsistency, which results in misaligned knowledge that distracts the distillation procedure. DKD outperforms F-L2 since it transfers the knowledge in logits, effectively mitigating the issue of position inconsistency in the intermediate features. However, the neglect of information within the intermediate features by DKD leads to insufficient transferred knowledge. For example, in the third example, DKD treats “clutter” (black) as “ceiling” (green). Meanwhile, the feature distillation methods, \textit{i.e.}, OFD and PEFD, achieve better results by effectively utilizing the rich information in the intermediate features. Compared to these competitors, our method achieves the best performance. As shown in Fig.~\ref{fig:sem}, our method excels in accurately segmenting both global and local semantic areas. For example, it successfully captures the global shape in the fifth example and accurately identifies the local object, such as sofa, in the second example. These results demonstrate the necessity of FMD in solving the position inconsistency problem, as well as the effectiveness of BKR in leveraging diverse knowledge within the intermediate features.

\begin{table*}
\renewcommand\arraystretch{1.2}
\centering
\caption{Ablation study on (a) shape classification, (b) object part segmentation, and (c) semantic segmentation. T and S represent the original teacher and the compressed student.}
\scalebox{0.95}{
\begin{tabular}{l|ccp{0.5cm}l|ccp{0.5cm}l|ccc}

\hline
\cline{1-12}
\multicolumn{3}{c}{\textbf{(a) Shape Classification}}&&\multicolumn{3}{c}{\textbf{(b) Object Part Segmentation}}&&\multicolumn{4}{c}{\textbf{(c) Semantic Segmentation}}\\

\cline{1-3}
\cline{5-7}
\cline{9-12}
Methods & OA & mAcc & &Methods & cat. mIoU & ins. mIoU && Methods & OA & mAcc & mIoU  \\

\cline{1-3}
\cline{5-7}
\cline{9-12}
(T)PointNet++ &92.54  &90.17 & & (T)PointNet++ &85.21  &81.74 & & (T)PointNet++ &82.75	&61.16	&52.23\\
(S)(1/8)PointNet++ &88.48  &80.61 & & (S)(1/8)PointNet++ &76.29  &58.05  & & (S)(1/8)PointNet++ &79.58	&57.25	&46.09 \\

\cline{1-3}
\cline{5-7}
\cline{9-12}
F-L2 &88.03  &79.69 & &F-L2 &75.92  &57.93  && F-L2 &79.26 &56.89	&45.34\\
REMD &88.86 &82.54 &&REMD &76.60 &58.05 &&REMD &79.78	&57.75	&47.01\\
FMD &89.06 &82.77 &&FMD &76.88 &58.14 &&FMD &80.23	&58.77	&47.66 \\
TDKR+FMD &89.75 &83.94 &&TDKR+FMD &77.46 &58.87 &&TDKR+FMD &80.47	&59.22	&48.64\\
BUKR+FMD &89.18 &83.09  &&BUKR+FMD  &77.20 &58.23 &&BUKR+FMD  &80.53	&58.89	&48.31\\
TDKR+BUKR+FMD & 89.92 & 84.43 &&TDKR+BUKR+FMD & 77.91 & 59.67 &&TDKR+BUKR+FMD & 80.80	&59.72	&49.41\\

\cline{1-3}
\cline{5-7}
\cline{9-12}
BKR+FMD (Ours)& \textbf{90.28} & \textbf{84.55} &&BKR+FMD (Ours)& \textbf{79.22} & \textbf{59.84}  &&BKR+FMD (Ours)& \textbf{81.02} & \textbf{60.30}  & \textbf{50.03} \\

\cline{1-12}
\hline
\end{tabular}\label{tab:ablation class}
}
\end{table*}

\begin{figure}
\setlength{\abovecaptionskip}{-0.2cm}
\setlength{\belowcaptionskip}{-0.4cm}
\centering
\includegraphics[width=4cm]{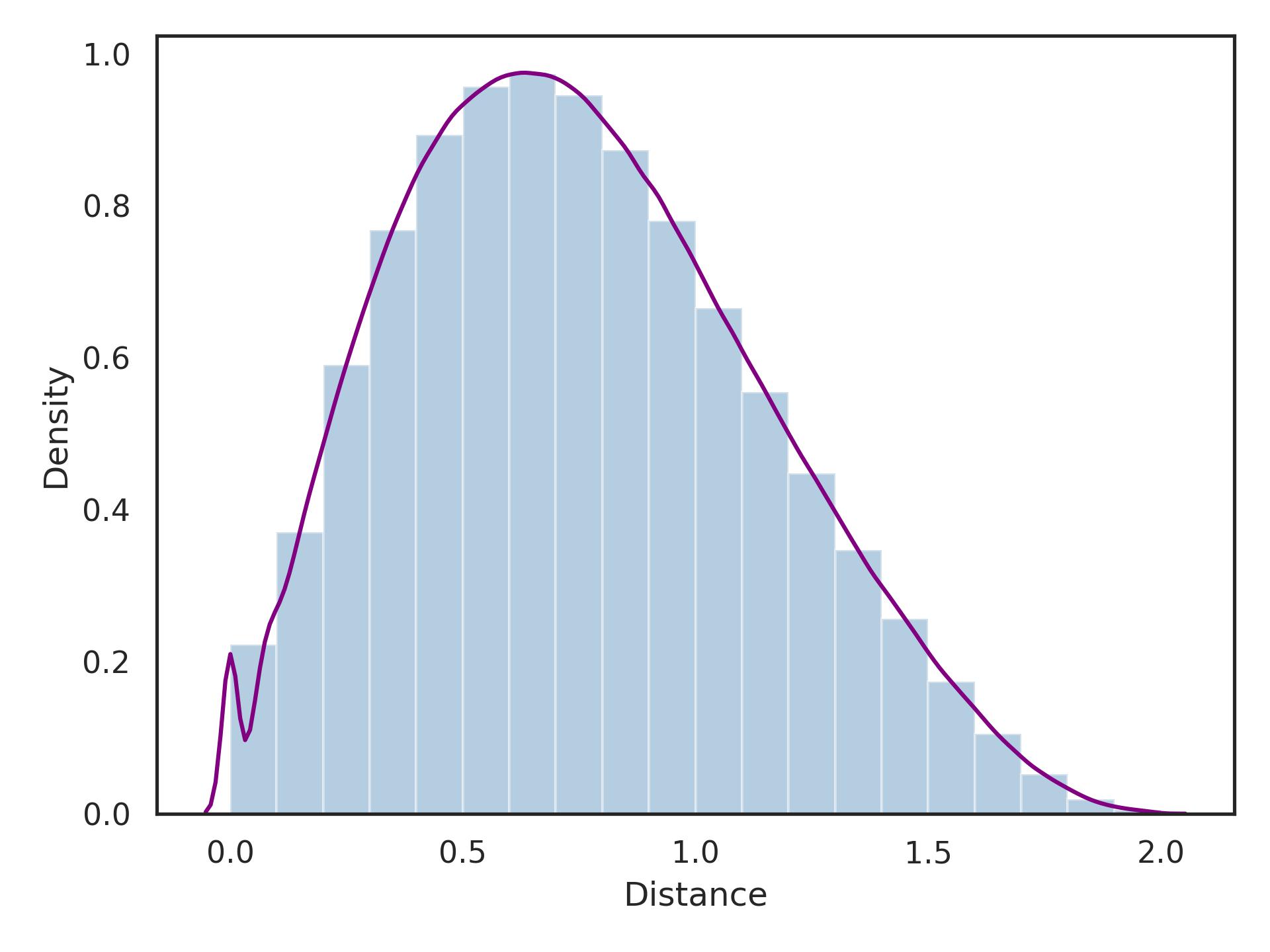}
\begin{center}
\caption{Histogram of the distance between point pairs.}\label{fig:inconsistency}
\end{center}
\end{figure}

\subsection{Analysis}
\subsubsection{Position Inconsistency in Feature Distillation}\label{sec:inconsistency}
To clarify the position inconsistency problem, we simulate the sampling process and visualize the normalized frequency histogram of the distance between point positions sampled by the teacher and the student.
In particular, we employ the preprocessed ModelNet40 dataset in which each input contains 1024 points and samples 512 points for the teacher and the student. The sampling process is the same as the first stage of many point cloud analysis models, such as PointNet++~\cite{qi2017pointnet++} and PointConv~\cite{wu2019pointconv}.
All the training data are used, and the distance between point pairs is calculated by Euclidean distance. We then count and plot the normalized frequency histogram.

\begin{table}
\scriptsize
\renewcommand\arraystretch{1.1}
\centering
\caption{Results of different feature levels.}
\setlength{\tabcolsep}{6mm}
\scalebox{0.9}{
\begin{tabular}{l|c|c}
\hline
\cline{1-3}
&(a) $L_2$ &(b) FMD\\
\hline
Levels &mIoU &  mIoU\\
\hline
Level-1 &46.18 & 47.14\\
\cline{1-3}
Level-2 &46.39 & 46.94\\
\cline{1-3}
Level-3 &47.74 & 47.98\\
\cline{1-3}
Level-4 &47.73 & 48.35\\
\cline{1-3}
Level-3,4  &46.52 & 48.55\\
\cline{1-3}
Level-2,3,4 &46.38 & 48.83\\
\cline{1-3}
Level-1,2,3,4 &45.34 & 49.19\\
\hline
\cline{1-3}
\end{tabular}
}
\label{tab:analysis}
\end{table}

As shown in Fig.~\ref{fig:inconsistency}, the distance is between $0\sim 2$ because the input point cloud positions are normalized to $-1\sim 1$ during data preprocessing.
Obviously, many point pairs have nonnegligible Euclidean distances. Specifically, approximately $26.98\%$ of the point pairs have a Euclidean distance greater than 1, indicating a large inconsistency between the teacher and student. Such inconsistency leads to misaligned intermediate features, which limits feature distillation effectiveness. There are also some point pairs with distances less than 0.25 or even equal to 0. These aligned or near-aligned points account for why other distillation methods can be effective without feature alignment.

We further analyze the influence of position inconsistency at different levels in feature distillation. 
We choose PointNet++ as the backbone and conduct experiments on the S3DIS dataset. Specifically, distillation is performed on the features of each level separately. 
Two distillation methods are employed.
One is our baseline which directly forces the student feature to mimic the teacher feature by $L_2$ loss. The other one replaces the distillation loss in the baseline with the proposed FMD.
The results are summarized in Table~\ref{tab:analysis}.
For L2, when distilling with single-level, it is observed that shallow features are more sensitive to position inconsistency than deep features. This is because the higher the feature level, the larger the perceptual area captured, which is able to represent more global shape information, making less misalignment between features of different positions. 
However, the performance is getting worse when more levels are added for distillation. This is because the utilization of more distillation levels may potentially lead to the accumulation of misaligned knowledge distillation. Besides, the inconsistency in knowledge transfer across different levels may further disrupt the distillation procedure, ultimately leading to a decline in performance. For FMD, since features are well aligned in FMD, both low-level and high-level knowledge can be well transferred from the teacher to the student, obtaining satisfactory results. Besides, as the shallow feature in Level-1 captures more local structure information which is significant in the semantic segmentation task, the better performance of Level-1 can be attributed to the abundant local structure information.
Besides, it is worth noting that employing multi-level features in feature distillation outperforms the single-level method, which is consistent with the observation of previous methods~\cite{chen2021cross,ji2021show}.

\begin{table}
\scriptsize
\renewcommand\arraystretch{1.4}
\centering
\caption{The universality and effectiveness of FMD. Experiments are conducted on (a) shape classification and (b) object part segmentation with PointNet++ as the backbone.}
\setlength{\tabcolsep}{0.8mm}
\scalebox{0.73}{
\begin{tabular}{l|ccccccc|c}

\hline
\cline{1-9}
\multicolumn{9}{c}{\textbf{(a) Shape Classification}}\\

\hline
& \textbf{F-L2}&\textbf{SP}&\textbf{FitNet}&\textbf{NST}&\textbf{AT}&\textbf{OFD} &\textbf{PEFD}&\textbf{BKR} \\

\hline
\textbf{OA}  &88.03 &88.31 &88.57 &88.61 &89.08 &89.23 &89.28 &89.68\\
\textbf{mAcc}  &79.69 &80.03 &82.17 &82.59 &82.79 &83.25 &83.27 &83.76\\

\hline
&\textbf{F-L2+FMD}&\textbf{SP+FMD}&\textbf{FitNet+FMD}&\textbf{NST+FMD}&\textbf{AT+FMD}&\textbf{OFD+FMD}&\textbf{PEFD+FMD}&\textbf{BKR+FMD}\\

\hline
\textbf{OA}  &88.81 &89.65 &88.85 &89.38 &89.47 &89.75 &89.68 &\textbf{90.28}\\
\textbf{mAcc}   &83.08 &83.66 &82.33 &83.87 &83.84 &83.93 &83.98 &\textbf{84.55}\\

\hline
\multicolumn{9}{c}{\textbf{(b) Object Part Segmentation}}\\

\hline
& \textbf{F-L2}&\textbf{SP}&\textbf{FitNet}&\textbf{NST}&\textbf{AT}&\textbf{OFD}&\textbf{PEFD}&\textbf{BKR}\\

\hline
\textbf{ins.mIoU} &75.92 &75.99 &76.34 &76.50 &76.71 &76.94 &77.25 &77.76\\
\textbf{cat.mIoU} &57.93 &57.97 &58.03 &58.37 &58.25 &58.60 &58.43 &58.44\\

\hline
&\textbf{F-L2+FMD}&\textbf{SP+FMD}&\textbf{FitNet+FMD}&\textbf{NST+FMD}&\textbf{AT+FMD}&\textbf{OFD+FMD}&\textbf{PEFD+FMD}&\textbf{BKR+FMD}\\
 
\hline
\textbf { ins.mIoU} &77.37 &77.78 &76.59 &77.61 &77.58 &77.76 &77.65 &\textbf{79.22}\\
\textbf{cat.mIoU}  &58.73 &58.59 &58.13 &58.41 &58.53 &58.79 &58.83 &\textbf{59.84}\\

\hline
\cline{1-9}
\end{tabular}
}
\label{tab:analysis2}
\end{table}

\subsubsection{Ablation Study }\label{sec:ablation}
To further demonstrate the effectiveness of the proposed BKR and FMD, 
we design an ablation study on the ModelNet40, ShapeNetPart, and S3DIS datasets with PointNet++ as the backbone. 
F-L2 is our baseline.
As shown in Table~\ref{tab:ablation class}, utilizing the proposed FMD can boost the performance.
In classification, FMD helps the student model outperform F-L2 by $3.08\%$.
The effectiveness of FMD is more remarkable on semantic segmentation. Specifically, FMD outperforms REMD by 0.45\%, 1.02\%, and 0.65\% on OA, mAcc, and mIoU, respectively.
We also combine FMD with other feature distillation methods. As shown in table~\ref{tab:analysis2}, FMD can consistently improve performance, demonstrating the universality and effectiveness of FMD.

In addition, we quantitatively analyze the effectiveness of TDKR, BUKR and BKR.
As shown in Table~\ref{tab:ablation class}, although BUKR+FMD improves the performance of FMD marginally, the collaboration effect between TDKR and BUKR is remarkable which is consistent with our conclusion that ``both local structure and global shape information are essential clues for point cloud''.
Taking object part segmentation as an example, on the one hand, TDKR+BUKR+FMD outperforms TDKR+FMD by 0.80\% on ins. mIoU, indicating the assisting role of BUKR to TDKR. On the other hand, TDKR+BUKR+FMD outperforms FMD by 1.03\% and 1.53\% on cat. mIoU and ins. mIoU, further proving the necessity of collaboration between TDKR and BUKR.
In addition, our framework (BKR+FMD), which combines TDKR, BUKR and residual connection, achieves the best results, showing that residual connections can bring rich information and improve knowledge transfer. 
Moreover, as shown in Table~\ref{tab:analysis2}, although the results of other feature distillation methods can be improved by FMD, they are still inferior to our method, \textit{i.e.}, BKR+FMD, further demonstrating the superiority of BKR.

\begin{table}
\scriptsize
\renewcommand\arraystretch{1.3}
\centering
\caption{Analysis of the parameter $k$.}
\scalebox{0.8}{
\setlength{\tabcolsep}{5mm}{
\begin{tabular}{l|ccccc}

\hline
\cline{1-6}
\multicolumn{6}{c}{\textbf{(a) Shape Classification}}\\

\hline
\textbf{k} &\textbf{1} &\textbf{3} &\textbf{5} &\textbf{7} &\textbf{9}\\

\hline
\textbf{OA}   &89.29 &89.49   &\textbf{90.28} &89.63 &89.60\\
\textbf{mAcc}   &83.84 &84.91  &\textbf{84.55} & 83.63 &83.29\\

\hline
\multicolumn{6}{c}{\textbf{(b) Object Part Segmentation}}\\

\hline
\textbf{k} &\textbf{1} &\textbf{3} &\textbf{5} &\textbf{7} &\textbf{9}\\

\hline
\textbf{ins.mIoU}  &76.72 &78.46  &\textbf{79.20} &78.92 &77.75\\
\textbf{cat.mIoU}  &58.14 &59.61  &\textbf{59.84} &59.40 &58.44\\

\hline
\cline{1-6}

\end{tabular}
}}
\label{tab:k}
\end{table}

\begin{table}
\scriptsize
\renewcommand\arraystretch{1.3}
\centering
\caption{Analysis of the tradeoff parameter $\lambda$.
}
\setlength{\tabcolsep}{5mm}

\scalebox{0.8}{
\begin{tabular}{l|ccccc}

\hline
\cline{1-6}
\multicolumn{6}{c}{\textbf{(a) Shape Classification}}\\

\hline
\textbf{$\lambda$} &\textbf{0.1} &\textbf{0.05} &\textbf{0.01} &\textbf{0.005} &\textbf{0.001}\\

\hline
\textbf{OA}   &\textbf{90.28} &89.91 &89.60 &89.82 &89.71\\
\textbf{mAcc}   &\textbf{84.55} &83.31  &83.76 &84.13 &83.28\\

\hline
\multicolumn{6}{c}{\textbf{(b) Object Part Segmentation}}\\

\hline
\textbf{$\lambda$} &\textbf{0.1} &\textbf{0.05} &\textbf{0.01} &\textbf{0.005} &\textbf{0.001}\\

\hline
\textbf{ins.mIoU}  &77.99 &78.41  &\textbf{79.20} &78.56 &78.23\\
\textbf{cat.mIoU}  &59.70 &59.29  &\textbf{59.84} &59.39 &59.22\\

\hline
\cline{1-6}

\end{tabular}
}
\label{tab:lambda}
\end{table}

\subsubsection{Analysis of Hyperparameters}\label{sec:param}
We analyze the nearest number $k$ in FMD and the tradeoff parameter $\lambda$ by cross-validation. Specifically, $20\%$ of the training set is used as the validation set, and the rest is employed to train the model. We vary $k$ in ${1,3, 5, 7, 9}$ and $\lambda$ in ${0.1, 0.05, 0.01, 0.005, 0.001}$. Experiments are conducted on ModelNet40~\cite{wu20153d} for shape classification and ShapeNetPart~\cite{yi2016scalable} for object part segmentation with PointNet++~\cite{qi2017pointnet++} as the backbone.

Table~\ref{tab:k} and Table~\ref{tab:lambda} present the results of varying $k$ and $\lambda$, respectively. Our method is more sensitive to the nearest number $k$ than the tradeoff parameter $\lambda$. This is because the number of nearest neighbors in FMD controls the receptive scale of the student and further determines the scope of contextual knowledge transferred from the teacher. Within a certain range, a larger $k$ will form a better representation of the local structure. After adding more neighbors, the performance will not increase because only a moderate $k$ can balance the local structure information and global shape knowledge. As shown in Table~\ref{tab:k}, we chose $k=5$ in our experiments.

\section{Conclusions}

In this paper, we design a universal feature distillation strategy for lightweight point cloud analysis. Since both the local structure knowledge and global shape knowledge of the teacher are essential for the student, a bidirectional knowledge reconfiguration (BKR) is presented to inherit the contextual knowledge from the teacher to all the scales of the student using bidirectional reconfiguration. Specifically, a top-down reconfiguration is developed for inheriting diverse local structure information, and a bottom-up reconfiguration is employed to inherit high-level shape knowledge. Since there exists a potential position inconsistency caused by the random point sampling operation in point cloud analysis, a feature mover's distance (FMD) is proposed to conduct the feature alignment. Experiments on shape classification, part segmentation and semantic segmentation benchmarks with various point cloud analysis networks show the effectiveness and 
universality of our framework. 

\section{Acknowledgments}
This research is sponsored by Beijing Nova Program (Z211100002121106), National Natural Science Foundation of China (Grant No. 62306041, 62276031).
\bibliographystyle{IEEEtran}
\bibliography{IEEEabrv,mylib}

\end{document}